\documentclass[10pt,twocolumn,letterpaper]{article}

 \usepackage{iccv}              % To produce the CAMERA-READY version
\definecolor{iccvblue}{rgb}{0.21,0.49,0.74}
\usepackage[pagebackref,breaklinks,colorlinks,allcolors=iccvblue]{hyperref}
\usepackage[misc]{ifsym}

\def\paperID{4317} % *** Enter the Paper ID here
\def\confName{ICCV}
\def\confYear{2025}

\title{F-ViTA: Foundation Model Guided Visible to Thermal Translation}

\author{Jay Nitin Paranjape\\
Johns Hopkins University\\
{\tt\small jparanj1@jhu.edu}
\and
Celso M de Melo\\
DEVCOM Army Research Laboratory\\
{\tt\small celso.m.demelo.civ@army.mil}
\and
Vishal M. Patel\\
Johns Hopkins University\\
{\tt\small vpatel36@jhu.edu}
}

\begin{document}
\maketitle
\begin{abstract}
Thermal imaging is crucial for scene understanding, particularly in low-light and nighttime conditions. However, collecting large thermal datasets is costly and labor-intensive due to the specialized equipment required for infrared image capture. To address this challenge, researchers have explored visible-to-thermal image translation. Most existing methods rely on Generative Adversarial Networks (GANs) or Diffusion Models (DMs), treating the task as a style transfer problem. As a result, these approaches attempt to learn both the modality distribution shift and underlying physical principles from limited training data.  In this paper, we propose F-ViTA, a novel approach that leverages the general world knowledge embedded in foundation models to guide the diffusion process for improved translation. Specifically, we condition an InstructPix2Pix Diffusion Model with zero-shot masks and labels from foundation models such as SAM and Grounded DINO. This allows the model to learn meaningful correlations between scene objects and their thermal signatures in infrared imagery. Extensive experiments on five public datasets demonstrate that F-ViTA outperforms state-of-the-art (SOTA) methods. Furthermore, our model generalizes well to out-of-distribution (OOD) scenarios and can generate Long-Wave Infrared (LWIR), Mid-Wave Infrared (MWIR), and Near-Infrared (NIR) translations from the same visible image. Code: \url{https://github.com/JayParanjape/F-ViTA/tree/master}.

%Thermal imaging plays an important role in scene understanding, especially in low-light and night-time conditions. However, it is expensive and tedious to create large datasets owing to specialized instruments required to capture a scene in the infrared domain, prompting researchers to study visible to thermal translation for images. Most existing methods employ Generative Adversarial Networks (GANs) or Diffusion Models (DMs) and approach this problem in a manner similar to style transfer.  Thus, they expect the model to inherently learn the distribution shift between the two modalities as well as the underlying physical principles from scratch, given the limited training data. In this paper, we instead propose utilizing the general world knowledge contained in foundation models to guide the diffusion process to produce better translations. More specifically, we condition an Instruct Pix2Pix Diffusion Model with zero-shot masks and labels from foundation models like SAM and Grounded DINO, allowing the model to learn correlations between an object in the scene and its thermal signature in the infrared image. Extensive experimentation shows that our method, called F-ViTA, outperforms various state-of-the-art (SOTA) methods on five public datasets. In addition, our model is able to perform well on Out-of-Distribution (OOD) translations and is able to generate Long-Wave (LWIR), Mid-Wave (MWIR) and Near Infrared (NIR) translations for the same visible image. Code will be released post-review.
\end{abstract}    
\begin{figure}
  \centering
  {\includegraphics[width=\linewidth]{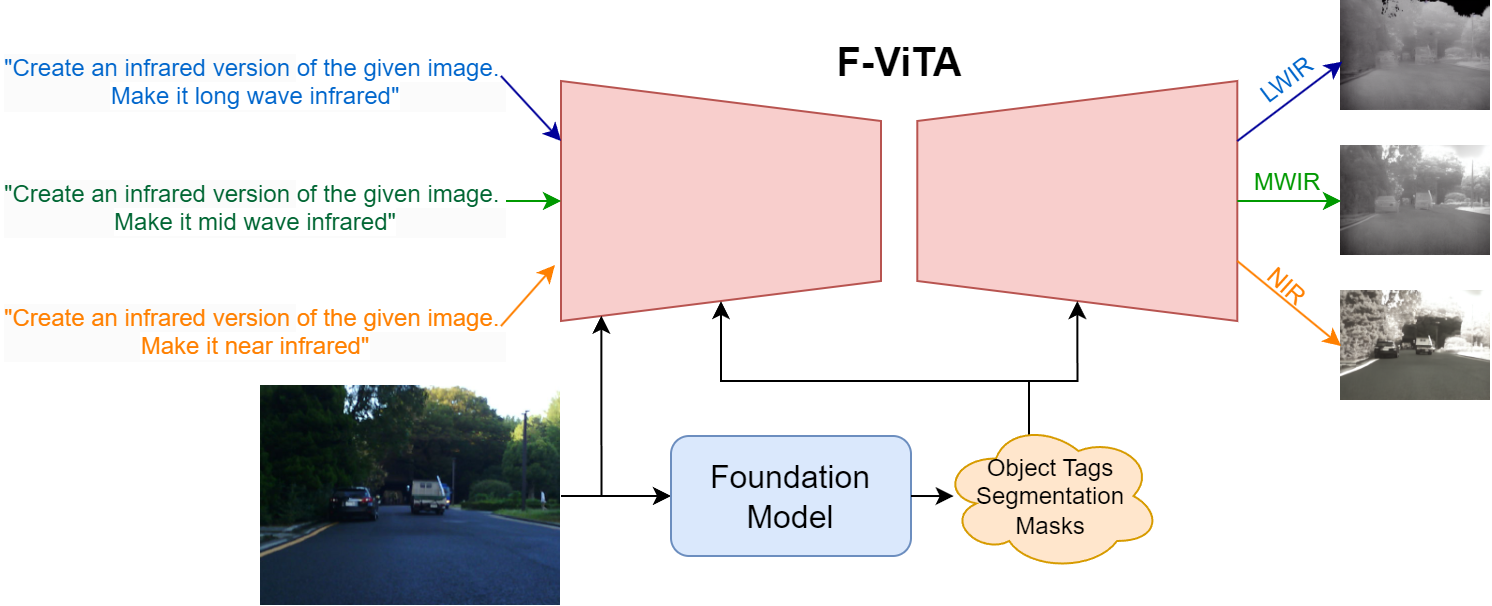}}
 \vskip-8pt 
 \caption{Our model, F-ViTA, leverages pretrained foundation models to extract object tags and segmentation masks from visible images in a zero-shot manner, using this information to enhance translation to the thermal domain. Additionally, F-ViTA enables user-guided infrared image generation through text prompts, allowing for the synthesis of specific infrared types--an ability not explored in existing methods.}  \label{fig:intro}
\end{figure}

\section{Introduction}
\label{sec:intro}

Many critical applications in autonomous driving, robotics, and surveillance require a precise understanding of natural scenes that remains robust in low-light and nighttime conditions. Thermal imaging has been widely explored for scene understanding and object recognition in such environments, where external illumination is impractical due to various constraints.

The infrared spectrum is typically divided into two regions: a reflection-dominated region, which includes the Near-Infrared (NIR) and Shortwave Infrared (SWIR) bands, and an emission-dominated thermal region, comprising the Midwave Infrared (MWIR) and Longwave Infrared (LWIR) bands. Given its resilience to challenging lighting conditions, thermal imagery provides a valuable source of information for these applications \cite{thermal_auto1,thermal_auto2,thermal_robotics1,thermal_robotics2,thermal_surveillance1,thermal_surveillance2}.

Recent studies have explored RGB-IR fusion to enhance tasks such as pedestrian detection \cite{pedestrian1,pedestrian2}, segmentation \cite{segmentation1,sigma}, and object detection \cite{objectdetection1,objectdetection2}. These approaches typically rely on paired RGB and thermal images from the same scene. However, capturing thermal images requires specialized hardware, including infrared cameras and sensors, making large-scale dataset collection both costly and labor-intensive. As a result, thermal datasets remain relatively scarce in the literature—especially paired RGB-thermal data—driving research efforts toward automatically generating infrared images from visible images.

Various methods have been proposed for image translation using Generative Adversarial Networks (GANs) \cite{eggan,infragan,thermalgan} and Diffusion Models (DMs) \cite{PID}. These approaches aim to learn the distribution shift between the two domains while preserving the physical properties of infrared images. However, in the low-data regime of RGB-to-thermal translation, such models are highly prone to overfitting, which significantly limits their generalizability.

Recently foundation models (FMs) have been introduced that are trained on large-scale natural data from the internet. These models learn highly representative features for visual data that are beneficial for various downstream tasks and promote generalized performance. Some examples include CLIP \cite{clip} for image-text alignment, Grounding DINO \cite{groundingdino} for open-world object detection, Segment Anything (SAM) \cite{sam} for prompted image segmentation and SAM2 \cite{sam2} for prompted video segmentation. Given the generalizable nature of such models, using the knowledge contained in them can help address the limited data issues with the current RBG-thermal translation methods. Motivated by this, we propose \textbf{F-ViTA}, which uses implicit guidance from FMs to improve the translation quality. For a given natural image, we first pass it to the FM to extract labeled masks of all objects present in the scene. Then, we inject this information during the diffusion process during training the translation model, which is based on Instruct Pix2Pix \cite{instructpix2pix}. This can encourage the model to correlate the output pixel intensity of the masked regions with the label of the object. For example, regions with the label ``person" should be given a higher pixel intensity in the thermal image. Such guidance also allows our model to do well on datasets with limited training pairs. Thus, our contributions are as follows:
\begin{enumerate}
    \item We present F-ViTA, a Diffusion Model-based approach for visible-to-thermal image translation that utilizes outputs from pretrained FMs to guide the translation process.
    \item We show that F-ViTA outperforms existing SOTA methods on five public datasets from NIR, MWIR and LWIR categories on multiple metrics.
    \item We also show that F-ViTA is able to generalize well to OOD data and can also generate the translated IR image from any of the three bands (LWIR, MWIR, NIR) given the user intent through a text prompt, as shown in \cref{fig:intro}. This is one of the first methods that shows such a functionality for visible-to-thermal translation. 
\end{enumerate}

\section{Related Work}
\label{sec:relatedwork}

Visible-to-thermal translation has been widely studied in the literature, with a majority methods attempting to solve this problem using GANs. Some early methods use Pix2Pix \cite{pix2pix2017}, an image-to-image translation model which uses conditional adversarial networks and cycle-GAN \cite{cyclegan}, which adds an additional consistency loss for stable training. These methods are not specific to visible-to-thermal translation. ThermalGAN \cite{thermalgan} was introduced for this task, with a focus on person identification using the thermal signature. InfraGAN \cite{infragan} is also a GAN-based approach that performs translation to the infrared domain for general scenes using pixel-wise loss functions and structural similarity. Recently, EGGAN \cite{eggan} was introduced that uses information from the edge map of the natural scene to guide the translation process. For this, they add an additional constraint to preserve the edges in the image.

\noindent As an alternative to the above GAN-based approaches, PID \cite{PID} was recently introduced, which uses DMs to perform the translation. In addition, this paper introduces physics-motivated loss functions that preserve the physical characteristics of the generated IR image. However, PID requires two-stage training, where the dataset-specific physics parameters are learnt first and then used as ground truths during the translation training. Furthermore, this work uses a common set of parameters for the dataset as a whole, rather than have a physical parameter for every object in the scene and has only been proposed for LWIR. Our method also uses a DM for translation, but it differs from PID in being a single-stage end-to-end approach. In addition, unlike PID, which explicitly learns dataset-level parameters, we implicitly try to learn object-level correlations by passing labeled maps during the diffusion process. Moreover, we show that our method produces good results for LWIR, MWIR as well as NIR images.

\section{Methodology}
\label{sec:method}
\subsection{Preliminaries: Grounded-SAM}
Given a visible image, we use Grounded-SAM \cite{groundedsam} to produce labeled masks. Grounded-SAM assembles Grounding-DINO \cite{groundingdino}, which is an open-world prompted object detector, and SAM \cite{sam}, which is a promptable image-segmentor. Given a text prompt and an image, Grounding DINO can detect the object corresponding to the text input and shows a strong zero-shot performance thanks to its training data. Similarly, SAM was trained on 11 million images and 1.1 billion masks and exhibits a strong zero-shot performance for segmenting out an object corresponding to a point, box or mask prompt. Thus, given a text descriptor of an object, Grounded-SAM first employs Grounding DINO to extract a bounding box for that object and passes it to SAM as a prompt, thus generating the labeled mask, where the input text is the label.

\begin{figure}[t!]
  \centering
  {\includegraphics[width=\linewidth]{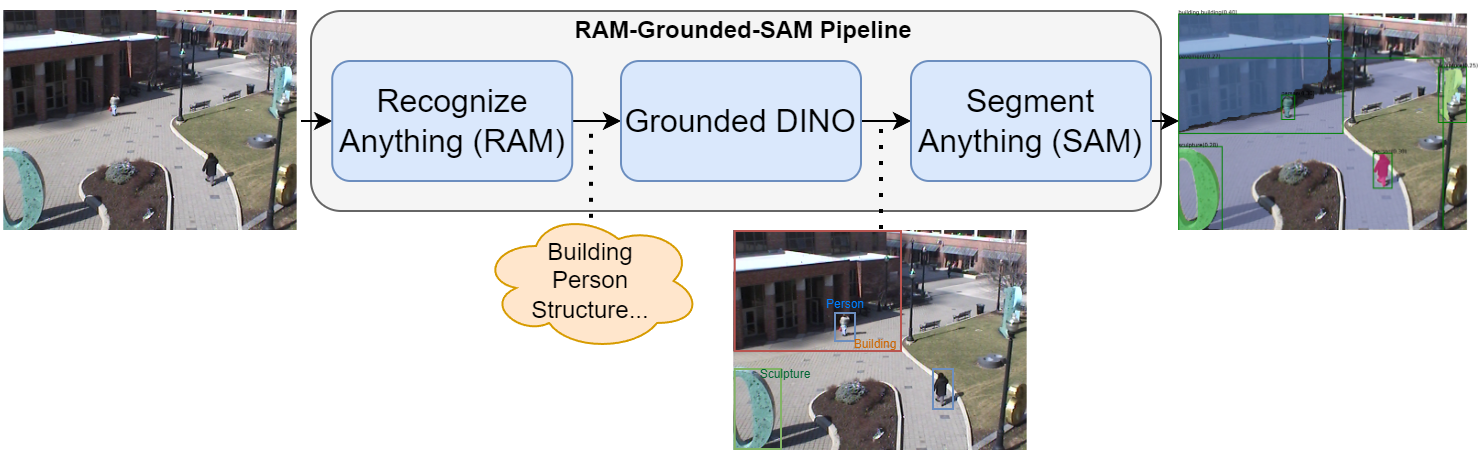}}
 \vskip-8pt 
 \caption{An example usage of the RAM-Grounded-SAM pipeline. The dotted lines indicate the output at every step.}  \label{fig:example_gsam}
\end{figure}

\noindent\textbf{RAM-Grounded-SAM: } Grounding-SAM can be used for automatic dense image annotation \cite{groundedsam}, where given an RGB image, the output is a list of masks for the objects in the scene, which are annotated with their corresponding text labels. For this, a foundational image tagging model like Recognize-Anything (RAM) \cite{ram} is used. RAM is trained to detect all objects in a scene and generate tags for them. It is trained on a large corpus of image-caption pairs from the internet, which are parsed to obtain the image with its corresponding tags. Given these pretrained foundation models, the RAM-Grounded-SAM pipeline passes an image through the RAM model to generate tags for all the present objects. These tags are passed to the Grounding-DINO model to generate corresponding bounding boxes, which are further passed to SAM to generate masks. Note that this entire pipeline does not involve any trainable parameters. \cref{fig:example_gsam} shows an example of the pipeline output. These foundation models are pretrained on a vast pool of diverse datapoints and thus are able to produce valuable priors for tasks like visible-to-infrared translation that are based in a low-data regime.

\begin{figure}[t!]
  \centering
  {\includegraphics[width=\linewidth]{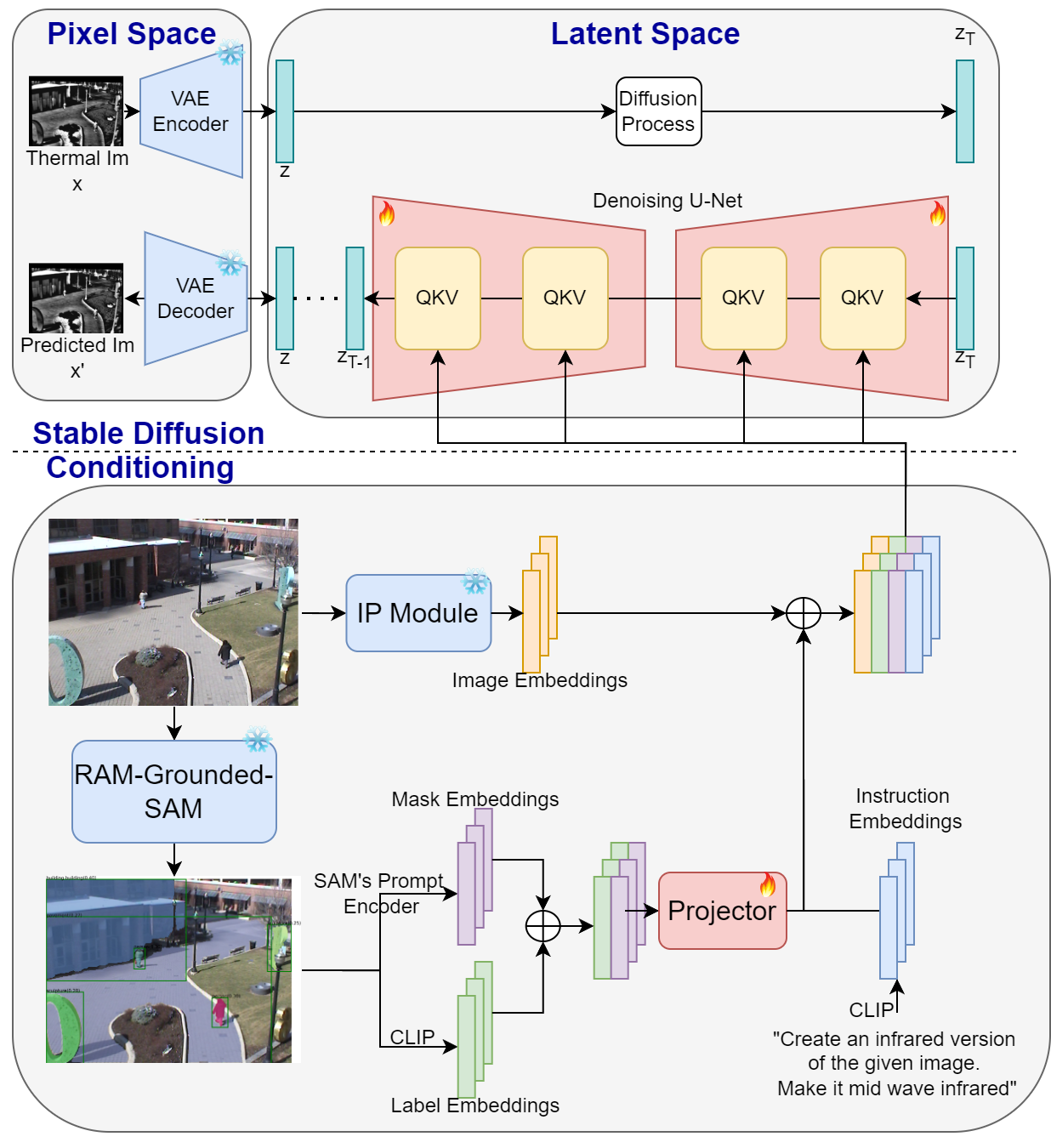}}
 \vskip-8pt 
 \caption{Training pipeline. The stable diffusion part is responsible for learning the distribution of thermal images. The conditioning part provides the visible image to enforce structural similarity as well as guidance from the foundation model and text instructions for improved translation. Only the Denoising UNet and the Projector module are trainable in the pipeline.}  \label{fig:arch}
\end{figure}

\subsection{Model Architecture}
We use Stable Diffusion \cite{stable_diffusion} for the translation process, similar to Instruct-Pix2Pix \cite{instructpix2pix}. As shown in \cref{fig:arch}, we use a pretrained encoder to convert the thermal image to the latent space, where it undergoes the forward diffusion process and the backward denoising process. The denoising is done using a UNet conditioned with additional prompts to control the generation process. The conditioning involves attention layers for each U-Net block, where the keys and the values are generated using the visible image and the text instructions and the queries are latent vectors at different timesteps. Finally, the pretrained decoder converts the generated latent back to the pixel space. In Instruct-Pix2Pix, the conditioning described before is done using an original image and a detailed text instruction, which describes the user-intended change to the original image. In contrast, F-ViTA also adds two additional conditioning embeddings derived from foundation model outputs. 

We pass the visible image through the frozen RAM-Grounded-SAM pipeline, which generates a number of text-annotated masks for the objects present in the image. We pass these masks through the prompt-encoder provided by SAM to generate discriminative embeddings. In addition, we pass the text-labels for every object through the pretrained CLIP encoder to generate text embeddings. These are concatenated along feature axis and passed through a trainable projector. This module is responsible for learning a transform that converts the FM embeddings to the space expected by Stable Diffusion. These are then concatenated to the image embeddings, obtained by passing the visible image through Instruct-Pix2Pix's IP module, and instruction embeddings, which are CLIP embeddings of the text instruction. The latter two embeddings are unique for a given image-instruction pair, while the former two vary for every object. Hence, we repeat the latter two embeddings to be equal to the number of detected objects, for concatenation. For the text instruction, we generate 50 synonymous sentences for the text: \textcolor{blue}{"Create an infrared version of the given image. Make it $<type>$ wave infrared."} Here, $<type>$ can take the values in $\{\text{near}, \text{mid}, \text{long}\}$. Thus, we generate a conditioning tensor using the image, information from the foundation models, and a text instruction. 

We train the entire pipeline on visible-thermal image pairs. Only the projector module and the denoising U-Net are the trainable modules in the pipeline. Since the conditioning tensors contain information about an object and its location in the image, we expect the model to learn physical correlations between the heat signatures of the object and the pixel intensity in the generated thermal image at the location of the object, for example, generating higher intensity pixels at the location of the object ``person". While existing methods expect the model to learn everything about the translation from the data, we utilize the guidance provided by FMs to reduce the load by providing localization and object information. Hence, our method is able to outperform existing methods in low-data regimes.
\section{Experiments and Results}

\subsection{Datasets}
\noindent \textbf{Long-Wave Infrared (LWIR): }We use the FLIR-ADAS \cite{FLIRADASDataset} and KAIST \cite{kaist} datasets within this wavelength range. FLIR-ADAS has thermal images in the range of $7.5-13.5 \mu m$. However, there is a slight misalignment between the natural and thermal images.  Therefore, we use the aligned version of this dataset \cite{FLIRalign}, which includes 4,890 training and 126 testing visible-thermal pairs. The KAIST dataset also contains thermal images within the $7.5$–$13.5 \mu m$ range, representing road scenes, with 12,538 training pairs and 2,252 testing pairs.

\noindent \textbf{Mid-Wave Infrared (MWIR): }We use two datasets to conduct experiments within this range. The OSU dataset \cite{osu} contains images from both the MWIR as well as the LWIR domain. There are 4,862 training pairs and 3,683 testing pairs. The LiTiV dataset \cite{litiv} also consists of scenes involving roads and pedestrians. It contains a total of 4,564 training and 1,761 testing pairs.

\noindent \textbf{Near Infrared (NIR): } We use the NIRScene \cite{nirscene} dataset to represent this wavelength, which contains sceneries and landscapes. This dataset is split into 381 training pairs and 96 testing pairs.

\subsection{Implementation Details}We use the SwinT-OGC checkpoint for Grounding DINO, swin large checkpoint for RAM and vit-base model for SAM. We retain the default settings of the foundation models during training as well as testing. The training was done on a single NVIDIA RTX6000 GPU with a batch size of 1. The base learning rate was set to 5e-5.
% \begin{table}
% \begin{center}
% \resizebox{\columnwidth}{!}{
% \begin{tabular}
% {@{\extracolsep{4pt}}c c c c c c c c c@{}}
% \toprule
%  & \multicolumn{4}{c}{FLIR-ADAS \cite{FLIRADASDataset}} & \multicolumn{4}{c}{KAIST \cite{kaist}} \\
% \cline{2-5} \cline{6-9} \\
% Method & FID (\(\downarrow\)) & LPIPS (\(\downarrow\)) & SSIM (\(\uparrow\)) & PSNR (\(\uparrow\)) & FID (\(\downarrow\)) & LPIPS (\(\downarrow\)) & SSIM (\(\uparrow\)) & PSNR (\(\uparrow\)) \\
% \midrule
% Pix2Pix \cite{pix2pix2017} & 199.06 & 0.38 & 0.30 & 16.34 & 132.04 & 0.22 & 0.69 & 21.25\\
% ThermalGAN \cite{thermalgan} & 407.01 & 0.60 & 0.15 & 11.59 & 277.85 & 0.24 & 0.66 & 19.74\\
% InfraGAN \cite{infragan} & 140.01 & 0.50 & \textbf{0.41} & 16.95 & 222.96 & 0.159 & 0.76 & \underline{22.97}\\
% EGGAN-M \cite{eggan} & 139.11 & 0.43 & 0.29 & 10.32 & 79.45 & 0.37 & 0.48 & 10.65\\
% EGGAN-U \cite{eggan} & 140.92 & 0.55 & 0.37 & 14.3 & 76.27 & 0.27 & 0.63 & 17.07\\
% % Instruct-Pix2Pix \cite{instructpix2pix} & 138.39 & 0.23 & 0.72 & 18.25 \\
% PID \cite{PID} & \textbf{84.26} & \underline{0.36} & 0.40 & \underline{17.26} & \textbf{51.69} & \underline{0.14} & 0.79 & \textbf{23.6}\\
% F-ViTA (Ours) & \underline{133.30} & \textbf{0.23} & \textbf{0.73} & \textbf{18.73} & \underline{76.18} & \underline{0.21} & \textbf{0.88} & 18.96\\
% \bottomrule
% \end{tabular}}
% \caption{Results on LWIR datasets. Bold - best, underline - second best}
% % \vskip-8pt
% \label{tab:lwir_results}
% \end{center}
% \end{table}

\begin{table}
\begin{center}
\resizebox{\columnwidth}{!}{
\begin{tabular}
{@{\extracolsep{4pt}}c c c c c@{}}
\toprule
Method & FID (\(\downarrow\)) & LPIPS (\(\downarrow\)) & SSIM (\(\uparrow\)) & PSNR (\(\uparrow\)) \\
\midrule
Pix2Pix \cite{pix2pix2017} & 199.06 & 0.38 & 0.30 & 16.34 \\
ThermalGAN \cite{thermalgan} & 407.01 & 0.60 & 0.15 & 11.59\\
InfraGAN \cite{infragan} & 140.01 & 0.50 & \underline{0.41} & 16.95\\
EGGAN-M \cite{eggan} & 139.11 & 0.43 & 0.29 & 10.32\\
EGGAN-U \cite{eggan} & 140.92 & 0.55 & 0.37 & 14.3\\
% Instruct-Pix2Pix \cite{instructpix2pix} & 138.39 & 0.23 & 0.72 & 18.25 \\
PID \cite{PID} & \textbf{84.26} & \underline{0.36} & 0.40 & \underline{17.26}\\
F-ViTA (Ours) & \underline{133.30} & \textbf{0.23} & \textbf{0.73} & \textbf{18.73}\\
\bottomrule
\end{tabular}}
\caption{Results on FLIR-ADAS \cite{FLIRADASDataset} dataset. Bold - best, underline - second best}
% \vskip-8pt
\label{tab:flir_results}
\end{center}
\end{table}

\begin{table}
\begin{center}
\resizebox{\columnwidth}{!}{
\begin{tabular}
{@{\extracolsep{4pt}}c c c c c@{}}
\toprule
Method & FID (\(\downarrow\)) & LPIPS (\(\downarrow\)) & SSIM (\(\uparrow\)) & PSNR (\(\uparrow\)) \\
\midrule
Pix2Pix \cite{pix2pix2017} & 132.04 & 0.22 & 0.69 & 21.25\\
ThermalGAN \cite{thermalgan} & 277.85 & 0.24 & 0.66 & 19.74\\
InfraGAN \cite{infragan} & 222.96 & 0.159 & 0.76 & \underline{22.97}\\
EGGAN-M \cite{eggan} & 79.45 & 0.37 & 0.48 & 10.65\\
EGGAN-U \cite{eggan} & 76.27 & 0.27 & 0.63 & 17.07\\
% Instruct-Pix2Pix \cite{instructpix2pix} & 138.39 & 0.23 & 0.72 & 18.25 \\
PID \cite{PID} & \textbf{51.69} & \textbf{0.14} & \underline{0.79} & \textbf{23.6}\\
F-ViTA (Ours) & \underline{76.18} & \underline{0.21} & \textbf{0.88} & 18.96\\
\bottomrule
\end{tabular}}
\caption{Results on KAIST dataset \cite{kaist}. Bold - best, underline - second best}
% \vskip-8pt
\label{tab:kaist_results}
\end{center}
\end{table}

\begin{table}
\begin{center}
\resizebox{\columnwidth}{!}{
\begin{tabular}
{@{\extracolsep{4pt}}c c c c c@{}}
\toprule
Method & FID (\(\downarrow\)) & LPIPS (\(\downarrow\)) & SSIM (\(\uparrow\)) & PSNR (\(\uparrow\)) \\
\midrule
EGGAN-M \cite{eggan} & 542.51 & 0.47 & 0.55 & 12.95\\
EGGAN-U \cite{eggan} & 75.76 & 0.18 & \textbf{0.86} & \underline{20.46}\\
% Instruct-Pix2Pix \cite{instructpix2pix} & 138.39 & 0.23 & 0.72 & 18.25 \\
PID \cite{PID} & 85.58 & 0.25 & 0.58 & 14.1\\
F-ViTA (Ours) & \textbf{61.97} & \textbf{0.15} & \underline{0.78} & \textbf{20.92}\\
\bottomrule
\end{tabular}}
\caption{Results on the OSU dataset \cite{osu}. Bold - best, underline - second best}
% \vskip-8pt
\label{tab:osu_results}
\end{center}
\end{table}

\begin{table}
\begin{center}
\resizebox{\columnwidth}{!}{
\begin{tabular}
{@{\extracolsep{4pt}}c c c c c@{}}
\toprule
Method & FID (\(\downarrow\)) & LPIPS (\(\downarrow\)) & SSIM (\(\uparrow\)) & PSNR (\(\uparrow\)) \\
\midrule
InfraGAN \cite{infragan} & 241.42 & 0.36 & 0.87 & 17.15\\
EGGAN-M \cite{eggan} & 238.41 & 0.35 & 0.93 & 12.81\\
EGGAN-U \cite{eggan} & \underline{152.09} & 0.23 & 0.95 & 18.45\\
% Instruct-Pix2Pix \cite{instructpix2pix} & 138.39 & 0.23 & 0.72 & 18.25 \\
PID \cite{PID} & 179.65 & \underline{0.20} & \underline{0.95} & \underline{19.22}\\
F-ViTA (Ours) & \textbf{118.52} & \textbf{0.16} & \textbf{0.96} & \textbf{27.55}\\
\bottomrule
\end{tabular}}
\caption{Results on the LiTiV2012 dataset \cite{litiv}. Bold - best, underline - second best}
% \vskip-8pt
\label{tab:litiv_results}
\end{center}
\end{table}

\begin{table}
\begin{center}
\resizebox{\columnwidth}{!}{
\begin{tabular}
{@{\extracolsep{4pt}}c c c c c@{}}
\toprule
Method & FID (\(\downarrow\)) & LPIPS (\(\downarrow\)) & SSIM (\(\uparrow\)) & PSNR (\(\uparrow\)) \\
\midrule
EGGAN-M \cite{eggan} & 126.2 & 0.18 & 0.83 & 15.25\\
EGGAN-U \cite{eggan} & 125.6 & 0.23 & 0.79 & 18.45\\
% Instruct-Pix2Pix \cite{instructpix2pix} & 138.39 & 0.23 & 0.72 & 18.25 \\
PID \cite{PID} & 166.67 & 0.31 & 0.62 & 14.21\\
F-ViTA (Ours) & \textbf{63.67} & \textbf{0.11} & \textbf{0.85} & \textbf{19.23}\\
\bottomrule
\end{tabular}}
\caption{Results on the NIRScene dataset \cite{nirscene}. Bold - best, underline - second best}
% \vskip-8pt
\label{tab:nirscene_results}
\end{center}
\end{table}

\subsection{Image Translation Experiments}
We compare our method with existing SOTA visible-to-thermal translators on five datasets. The results on FLIR-ADAS are tabulated in \cref{tab:flir_results}. We find that our method is able to outperform existing methods on multiple metrics, with a significant rise observed in SSIM. However, we see a lower FID score than the current SOTA. While FID compares the distributions between the ground truth and the generated thermal image, SSIM measures the structural and visual soundness of the generated image. Hence, both are important to gauge the quality of generation. On the KAIST dataset, as can be seen in \cref{tab:kaist_results}, we see a similar trend where the SSIM for our method is better but other metrics are comparable to PID. This is not surprising, given that PID models the physical quantities behind LWIR images and constrains the model accordingly. In contrast, our model captures the structure of objects better due to localization information present through the masks (causing improved SSIM), but at the same time has a slight tendency to miss the differences in output intensities for different parts of the same object (causing increased FID). This can happen since there are multiple similar signals from the prompts for different parts of the same object. In addition, note that FLIR is a comparatively smaller dataset than KAIST, which is why we see an overall better performance with our method.

\begin{figure*}
  \centering
  {\includegraphics[width=0.85\linewidth]{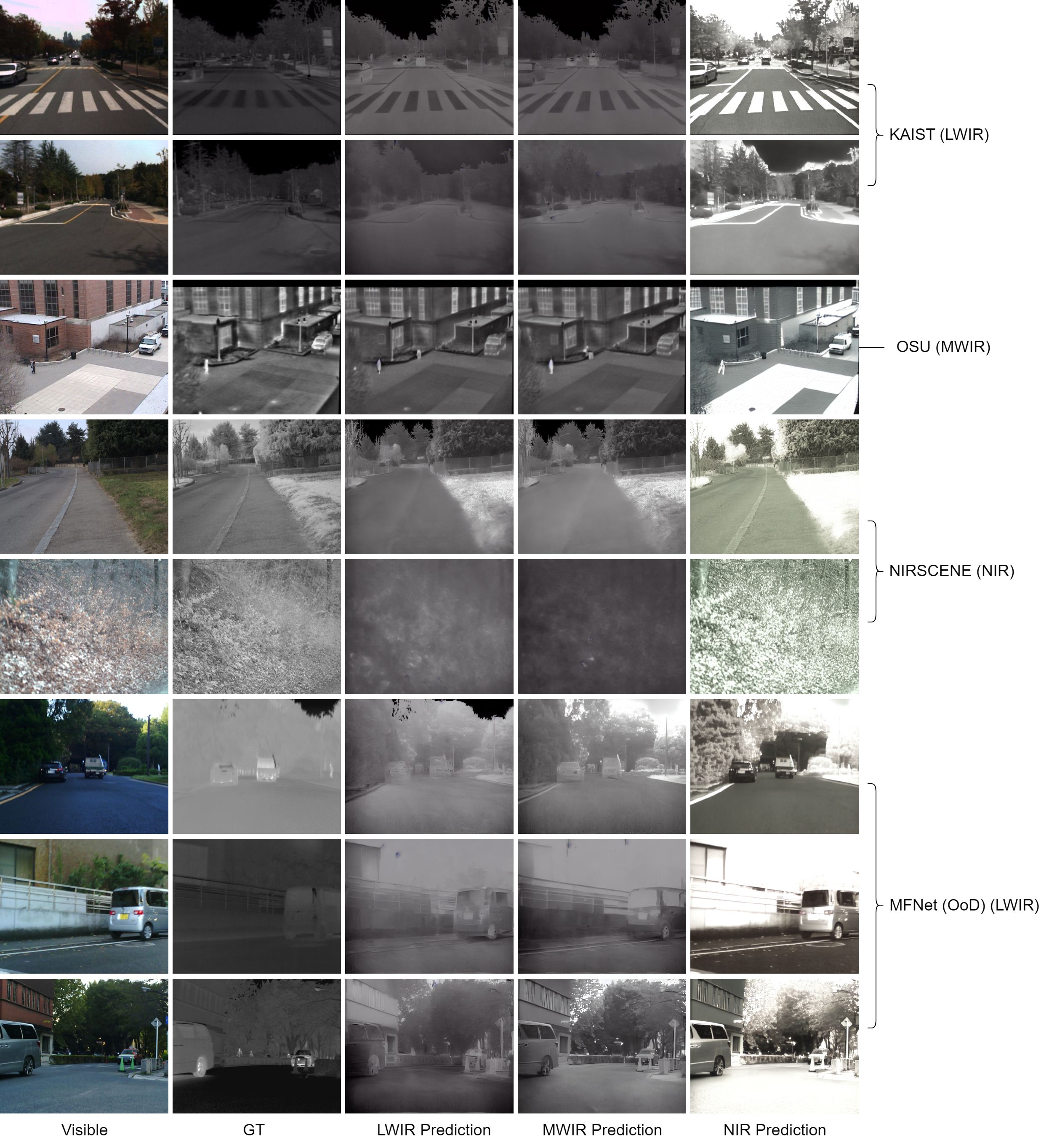}}
 \vskip-8pt 
 \caption{Qualitative Results over datasets from three different wavelength spectra. Our method (third column) is able to generate images more similar to the ground truth as compared to existing methods (fourth, fifth and sixth columns)}  \label{fig:qual}
\end{figure*}

\cref{tab:osu_results} shows the results on the OSU dataset, where we see a significant improvement in all metrics, while SSIM is comparable. Similarly, for LiTiV, as shown in \cref{tab:litiv_results}, and for NIRScene, as seen in \cref{tab:nirscene_results}, we find that our method outperforms all other methods across all metrics.  Note that these are cases of low-data regimes, where the advantages of our model are clearly demonstrated. We present qualitative results in \cref{fig:qual} across all five datasets used for comparison, selecting examples from both indoor and outdoor scenes. As shown in the figure, our method (third column) performs consistently well across all datasets. PID excels in LWIR but struggles to generate accurate translations for MWIR and shows an overall increased intensity in the NIR case. This is expected, as the theoretical analysis and formulation of PID are primarily designed for LWIR translations. In contrast, EGGAN-M and EGGAN-U slightly distort object structures, as observed in the figure. In comparison, our method, F-ViTA, generates thermal images that closely resemble the ground truth.

\subsection{Out-of-Distribution (OOD) Experiments}
%We use the MFNet \cite{mfnet} dataset for evaluating the generalizability of our method on the OOD data. MFNet contains visible images and their thermal counterparts in the LWIR range. As shown in \cref{tab:mfnet_ood_results}, F-ViTA, trained on the KAIST dataset is able to perform well on MFNet as compared to PID trained on KAIST. Similarly, for models tuned using only FLIR, our method performs significantly better than PID trained on the same dataset. PID first learns estimates for physical quantities related to the thermal dataset as a whole, while our method learns object-wise estimates implicitly. Hence, when transferred to a new dataset, our approach shows more robustness, which can be seen through the table. In addition, we also include one more experiment where we train F-ViTA using one dataset from each modality (KAIST for LWIR, LiTiV for MWIR and NIRScene or NIR) and then test its performance on MFNet with the instruction of converting it to a long wave infrared. We observe that even after the addition of additional modalities, the performance on MFNet remains comparable to when trained using just the LWIR KAIST dataset.

We use the MFNet \cite{mfnet} dataset to evaluate the generalizability of our method on out-of-distribution (OOD) data. MFNet contains visible images and their thermal counterparts in the LWIR range. As shown in \cref{tab:mfnet_ood_results}, F-ViTA, trained on the KAIST dataset, performs significantly better on MFNet compared to PID trained on KAIST. Similarly, for models tuned using only the FLIR dataset, our method outperforms PID trained on the same data. PID first learns estimates for physical quantities related to the thermal dataset as a whole, while our method implicitly learns object-wise estimates. As a result, our approach exhibits greater robustness when transferred to a new dataset, as reflected in the table. Additionally, we conduct an experiment where we train F-ViTA using one dataset from each modality (KAIST for LWIR, LiTiV for MWIR, and NIRScene or NIR for NIR) and then test its performance on MFNet with the instruction to convert to long-wave infrared. We find that even with the inclusion of additional modalities, the performance on MFNet remains comparable to when trained using only the LWIR KAIST dataset.

\begin{table}
\begin{center}
\resizebox{\columnwidth}{!}{
\begin{tabular}
{@{\extracolsep{4pt}}c c c c c@{}}
\toprule
Method & FID (\(\downarrow\)) & LPIPS (\(\downarrow\)) & SSIM (\(\uparrow\)) & PSNR (\(\uparrow\)) \\
\midrule
PID \cite{PID} (tuned on KAIST) & 196.10 & 0.39 & 0.85 & 13.53\\
Ours (tuned on KAIST) & \textbf{70.98} & \textbf{0.30} & \textbf{0.87} & \textbf{16.33}\\
\midrule
PID \cite{PID} (tuned on FLIR-ADAS) & 142.60 & 0.47 & 0.70 & 12.02\\
Ours (tuned on FLIR-ADAS) & \textbf{100.54} & \textbf{0.35} & \textbf{0.81} & \textbf{12.85}\\
\midrule
Ours (tuned on all datasets from all wavelengths) & 94.5 & 0.37 & 0.83 & 13.72\\
\bottomrule
\end{tabular}}
\caption{OOD Results on the MFNet dataset \cite{mfnet}. Bold - best, underline - second best}
% \vskip-8pt
\label{tab:mfnet_ood_results}
\end{center}
\end{table}

\subsection{Text-Prompted Translation Experiments}
\begin{figure*}
  \centering
  {\includegraphics[width=0.8\linewidth]{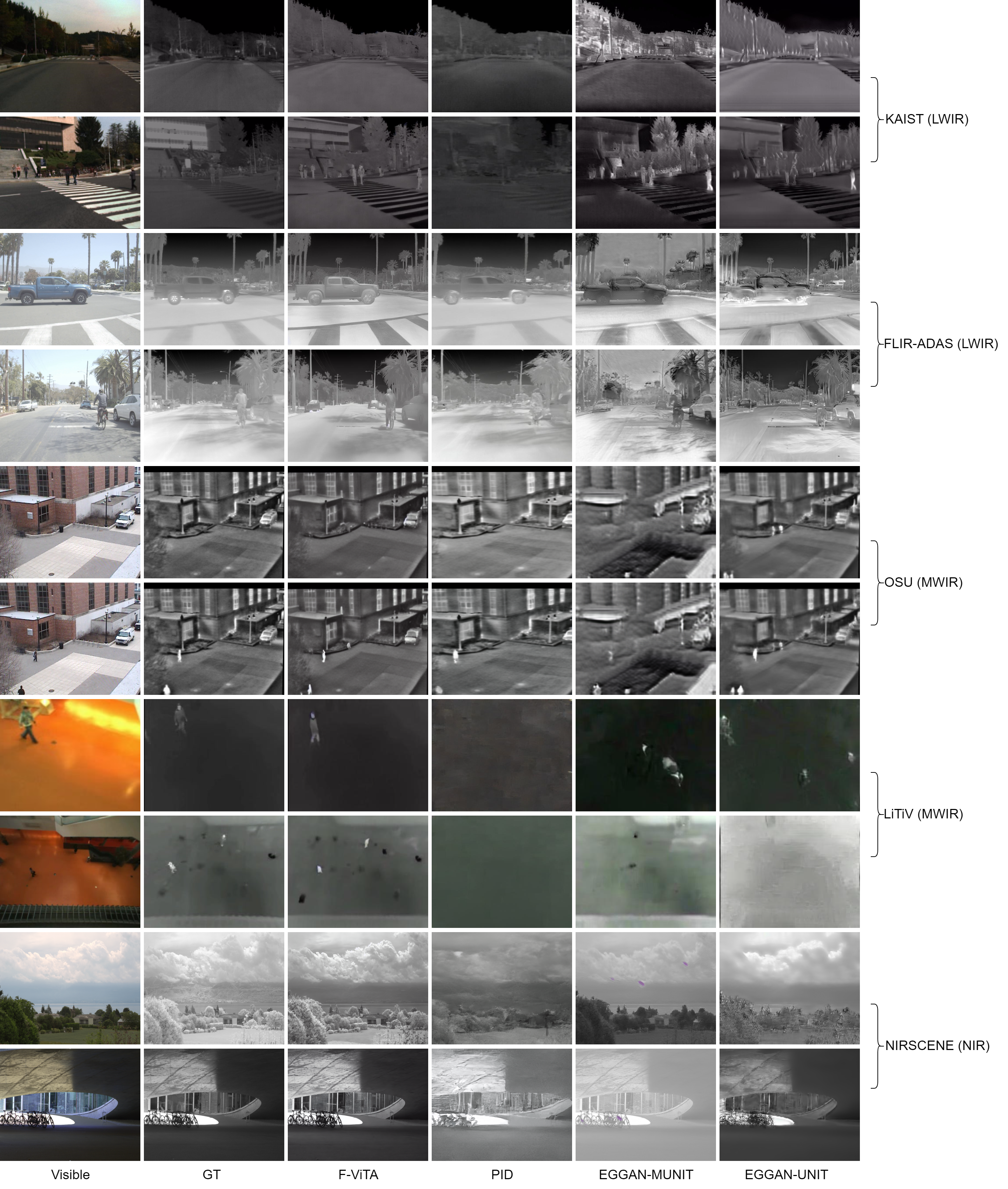}}
 \vskip-8pt 
 \caption{Text prompted translation capability of F-ViTA. Our method is able to generate LWIR, MWIR or NIR images based on the text instruction. The second column shows the ground truth which is from the wavelength range specified next to the dataset name on the right.}  \label{fig:allmodal_translations}
\end{figure*}

F-ViTA uses text instructions for guiding the diffusion model to perform the image translation. Hence, it is possible for it to learn to generate images for specific infrared spectra by specifying it in the instructions. To explore this, we train F-ViTA on KAIST (LWIR), OSU (MWIR), LiTiV (MWIR) and NIRSCENE (NIR) together with the type of image in the instruction. For instance, for KAIST, we ask the DM to generate a long wave infrared image. Once trained, we test the model's ability to generate all three types of IR images from a given RGB image from the test data. Some examples of such translations can be seen in \cref{fig:allmodal_translations}.
From the figure, we observe that F-ViTA produces reasonable predictions for LWIR, MWIR, and NIR given a single RGB image. The GT column represents the ground truth thermal image from the dataset-specific wavelength range. The first five rows display images from the test sets of the datasets on which the model was trained. The last three rows show predictions on MFNet, which was not used during training, demonstrating strong out-of-distribution (OOD) performance with F-ViTA. The final row illustrates a failure case where the model misses pedestrians in the MWIR and NIR predictions, indicating potential areas for improvement. Notably, no existing methods demonstrate such multi-spectral translation capability. We are the first to explore this direction with F-ViTA and encourage further research in this area.

\subsection{Downstream Applications}

\begin{table}
\begin{center}
\resizebox{0.85\columnwidth}{!}{
\begin{tabular}
{@{\extracolsep{4pt}}c c c@{}}
\toprule
$\%$ original data used & MIoU (\(\uparrow\)) & Mean Pixel Accuracy (\(\uparrow\)) \\
\midrule
0 & 0.501 & 55.58\\
10 & 0.503 & 56.01\\
25 & 0.520 & 59.22\\
50 & 0.526 & 59.1\\
100 & 0.527 & 59.6\\
\bottomrule
\end{tabular}}
\caption{Results on MFNet \cite{mfnet} using Sigma \cite{sigma} as the RGB-T segmentation pipeline. The rows indicate the percentage of real thermal data used in the training data, with the rest of the data being synthetic. We see only a small drop in the metrics as we go up the table (more synthetic data used), towards the applicability of our method.}
% \vskip-8pt
\label{tab:rgbt_results}
\end{center}
\end{table}

\noindent \textbf{RGB-T Fusion for Segmentation: } Thermal imagery is often used for multi-modal segmentation, where the RGB image and its paired thermal image are processed and fused to produce the output segmentation map. We conduct experiments to show an application of our pipeline for thermal image generation. We use the F-VITA model, tuned on the KAIST dataset, to generate thermal images for the MFNet \cite{mfnet} dataset. Next, we generate five types of data for training a RGB-T segmentation model, using $0\%$, $10\%$, $25\%$, $50\%$, and $100\%$ of the actual thermal images respectively. The rest of the images are taken from the generated predictions from our model. We use the original thermal images for the testing data. We use the SOTA RGB-T fusion model, called Sigma \cite{sigma} for the segmentation task. Sigma uses the recently introduced V-Mamba framework along with a concatenation and cross-attention mechansim to effectively fuse the two modalities. With this setup, we train Sigma with the five datasets and test its performance, as tabulated in \cref{tab:rgbt_results}.

\noindent We observe that when only the original thermal data is used, Sigma gives a MIoU of $52.7$ and mean pixel accuracy of $59.6$. When we replace $50\%$ of the data with the synthetic data generated with our method, the MIoU drops to $52.6$ and accuracy to $59.1$, which is not a significant drop compared to the reduction in original training data. With an all-synthetic training set, these numbers only reduce by 4 points. Note that the thermal data was created using F-ViTA trained on KAIST only, thus being OOD for the generation model. Thus, these results also denote the quality of our predictions.

\begin{table}
\begin{center}
\resizebox{0.8\columnwidth}{!}{
\begin{tabular}
{@{\extracolsep{4pt}}c c c@{}}
\toprule
$\%$ original data used & MIoU (\(\uparrow\)) & Mean Pixel Accuracy (\(\uparrow\)) \\
\midrule
0 & 0.30 & 43.38\\
10 & 0.30 & 45.24\\
25 & 0.30 & 40.01\\
50 & 0.38 & 51.29\\
100 & 0.38 & 54.12\\
\bottomrule
\end{tabular}}
\caption{Results on MFNet \cite{mfnet} using Segformer \cite{segformer} for thermal segmentation for 10 classes. The rows indicate the percentage of real thermal data used in the training data, with the rest of the data being synthetic. We see only a slight drop in the metrics as we go up the table (more synthetic data used), which suggests a use case of F-ViTA for synthetic data generation.}
% \vskip-8pt
\label{tab:thermal_results}
\end{center}
\end{table}

\noindent \textbf{Thermal Image Segmentation: }In this experiment, we train a Segformer \cite{segformer} on the thermal training data using the same mixtures as in the previous experiment, and then evaluate its performance on the test data containing original thermal images. Note that we do not use RGB images here, making this a purely semantic segmentation task based on thermal images. As shown in \cref{tab:thermal_results}, with 100\% original training data, we achieve a MIoU of 0.38 and a mean accuracy of 54\%. Even when replacing 50\% of the data with synthetic images, we observe comparable performance. Below this threshold, we see a noticeable decrease in accuracy, suggesting areas for further exploration. Nevertheless, the predictions from our method offer a viable alternative to capturing infrared images using specialized hardware.

\begin{table}
\begin{center}
\resizebox{0.7\columnwidth}{!}{
\begin{tabular}
{@{\extracolsep{4pt}}c c c c@{}}
\toprule
$\%$ original data used & P (\(\uparrow\)) & R (\(\uparrow\)) & MAP@50 (\(\uparrow\))\\
\midrule
0 & 0.63 & 0.45 & 0.48\\
10 & 0.64 & 0.56 & 0.57\\
25 & 0.64 & 0.60 & 0.60\\
50 & 0.67 & 0.62 & 0.62\\
100 & 0.67 & 0.59 & 0.62\\
\bottomrule
\end{tabular}}
\caption{Results on KAIST \cite{kaist} using YOLOv8 \cite{yolov8_ultralytics} for pedestrian detection. The rows indicate the percentage of real thermal data used in the training data, with the rest of the data being synthetic. We see only a slight drop in the metrics as we go up the table (more synthetic data used), which suggests a use case of F-ViTA for synthetic data generation.}
% \vskip-8pt
\label{tab:pedestrian_results}
\end{center}
\end{table}
\noindent \textbf{Pedestrian Detection in Thermal Imagery: }In this experiment, we use the thermal images generated by F-ViTA for the KAIST test dataset as synthetic data to generate five splits, with $0\%$, $10\%$, $25\%$, $50\%$ and $100\%$ being the real data percentage and the rest filled using synthetic data as done previously. Then, we train YOLOv8 \cite{yolov8_ultralytics} on these splits and test the performance on real thermal data, which is not used during training. As shown in \cref{tab:pedestrian_results}, we calculate the precision (P), recall (R) and the mean average precision at 0.5 threshold (mAP@50) for all cases. We observe that even after replacing $50\%$ of the real data, there is on par performance and surprisingly, a higher recall. As we go up the table, we see that the drop in performance is quite small compared to the amount of real data replaced, thus signifying the usefulness of the generated thermal images for such downstream tasks. Similar to the segmentation task, while we see that completely using the thermal data still has room for improvement, F-ViTA offers a viable solution for semi-supervised methods as seen from the table.

\subsection{Ablation Studies}
We perform an ablation study over the components extracted from the foundation models that are passed to the diffusion process. As seen in \cref{tab:ablation_components}, when neither of text, bounding boxes or mask embeddings are used (first row), we see a lower performance in all metrics. Adding just the text tags identified from the image (second row) improves these metrics. The text tags are extracted using the RAM model, and converted into embeddings using CLIP before passing to the DM. Next, we add only the embeddings from the bounding boxes for the objects detected by Grounding DINO (third row) and see on par metric values. However, adding mask embeddings from Grounded SAM (fourth row) gives us a slight improvement. This is expected since masks provide more precise localization than bounding boxes. In addition, unlike masks, a  bounding box around bigger objects like buildings or trees may also encompass other objects like vehicles or people. Hence, we see an advantage when using masks. In the fifth row, we use both the text tags and the bounding boxes, which slightly improves the performance over second and third rows, where they are passed separately. Finally, the last row denotes the case used in F-ViTA, with both the text embeddings and mask embeddings being provided to the DM. We see that this case improves over all other rows, since text and masks provide the localization and semantic information about the objects to the DM. We do not include cases where box and mask embeddings are used together, since adding boxes is superfluous in the presence of masks.
\begin{table}
\begin{center}
\resizebox{\columnwidth}{!}{
\begin{tabular}
{@{\extracolsep{4pt}}c c c c c c c@{}}
\toprule
Text Labels & Bounding Boxes & Masks & FID (\(\downarrow\)) & LPIPS (\(\downarrow\)) & SSIM (\(\uparrow\)) & PSNR (\(\uparrow\)) \\
\midrule
& & & 80.03& 0.18 & 0.72 & 19.64\\
\checkmark & & & 76.43 & 0.20 & 0.74 & 20.5\\
 & \checkmark & & 76.01 & 0.20 & 0.75 & 20.01\\
& & \checkmark & 69.81 & 0.19 & 0.76 & 20.7\\
\checkmark & \checkmark & & 75.18 & 0.17 & 0.76 & 20.24\\
\checkmark & & \checkmark & \textbf{61.97} & \textbf{0.15} & \textbf{0.78} & \textbf{20.92}\\
\bottomrule
\end{tabular}}
\caption{Ablation over the components passed to the diffusion process using the FM. \checkmark indicates the component being used, with the first row indicating neither text, boxes or masks being used and the last row indicating F-ViTA where text and masks are used.}
% \vskip-8pt
\label{tab:ablation_components}
\end{center}
\end{table}
\section{Conclusion}

In this work, we introduce F-ViTA, a diffusion model-based visible-to-thermal image translator that leverages pretrained foundation models to guide the diffusion process. We demonstrate that incorporating object identification, location, and semantic information via segmentation masks and text labels significantly enhances translation, particularly in low-data regimes. Extensive experimentation on five public datasets across LWIR, MWIR, and NIR spectra shows that F-ViTA outperforms or matches existing methods on multiple metrics, with improved generalization on out-of-distribution datasets. Notably, F-ViTA is the first method to enable conversion of the same RGB image into LWIR, MWIR, or NIR based on user-provided text instructions, opening new avenues for future research. We also highlight the potential applications of visible-to-thermal translation in downstream tasks like RGB-T fusion and thermal segmentation, demonstrating its value for semi-supervised learning in data-limited scenarios. Future work should further explore text-prompted inter-modality transfer, with applications in biometrics, robotics, and surveillance.

{
    \small
    \bibliographystyle{ieeenat_fullname}
    \bibliography{main}

\begin{thebibliography}{35}
\providecommand{\natexlab}[1]{#1}
\providecommand{\url}[1]{\texttt{#1}}
\expandafter\ifx\csname urlstyle\endcsname\relax
  \providecommand{\doi}[1]{doi: #1}\else
  \providecommand{\doi}{doi: \begingroup \urlstyle{rm}\Url}\fi

\bibitem[Brooks et~al.(2023)Brooks, Holynski, and Efros]{instructpix2pix}
Tim Brooks, Aleksander Holynski, and Alexei~A Efros.
\newblock Instructpix2pix: Learning to follow image editing instructions.
\newblock In \emph{Proceedings of the IEEE/CVF Conference on Computer Vision and Pattern Recognition}, pages 18392--18402, 2023.

\bibitem[Brown and S\"usstrunk(2011)]{nirscene}
M. Brown and S. S\"usstrunk.
\newblock Multispectral {SIFT} for scene category recognition.
\newblock In \emph{Computer Vision and Pattern Recognition (CVPR11)}, pages 177--184, Colorado Springs, 2011.

\bibitem[Castro~Jimenez and Mart{\'\i}nez-Garc{\'\i}a(2016)]{thermal_robotics1}
Lidice~E Castro~Jimenez and Edgar~A Mart{\'\i}nez-Garc{\'\i}a.
\newblock Thermal image sensing model for robotic planning and search.
\newblock \emph{Sensors}, 16\penalty0 (8):\penalty0 1253, 2016.

\bibitem[Dai et~al.(2021)Dai, Yuan, and Wei]{thermal_auto2}
Xuerui Dai, Xue Yuan, and Xueye Wei.
\newblock Tirnet: Object detection in thermal infrared images for autonomous driving.
\newblock \emph{Applied Intelligence}, 51\penalty0 (3):\penalty0 1244--1261, 2021.

\bibitem[Davis and Sharma(2007)]{osu}
James~W. Davis and Vinay Sharma.
\newblock Background-subtraction using contour-based fusion of thermal and visible imagery.
\newblock \emph{Computer Vision and Image Understanding}, 106\penalty0 (2):\penalty0 162--182, 2007.
\newblock Special issue on Advances in Vision Algorithms and Systems beyond the Visible Spectrum.

\bibitem[Deng et~al.(2021)Deng, Tian, Huang, Xiong, and Bi]{pedestrian1}
Qing Deng, Wei Tian, Yuyao Huang, Lu Xiong, and Xin Bi.
\newblock Pedestrian detection by fusion of rgb and infrared images in low-light environment.
\newblock In \emph{2021 IEEE 24th International Conference on Information Fusion (FUSION)}, pages 1--8. IEEE, 2021.

\bibitem[Filippini et~al.(2020)Filippini, Perpetuini, Cardone, Chiarelli, and Merla]{thermal_robotics2}
Chiara Filippini, David Perpetuini, Daniela Cardone, Antonio~Maria Chiarelli, and Arcangelo Merla.
\newblock Thermal infrared imaging-based affective computing and its application to facilitate human robot interaction: A review.
\newblock \emph{Applied Sciences}, 10\penalty0 (8):\penalty0 2924, 2020.

\bibitem[{FLIR Systems}(2025)]{FLIRADASDataset}
{FLIR Systems}.
\newblock Flir adas dataset.
\newblock \url{https://www.flir.com/oem/adas/adas-dataset-form/}, 2025.
\newblock Accessed: 2025-02-20.

\bibitem[Hwang et~al.(2015)Hwang, Park, Kim, Choi, and Kweon]{kaist}
Soonmin Hwang, Jaesik Park, Namil Kim, Yukyung Choi, and In~So Kweon.
\newblock Multispectral pedestrian detection: Benchmark dataset and baseline.
\newblock In \emph{2015 IEEE Conference on Computer Vision and Pattern Recognition (CVPR)}, pages 1037--1045, 2015.

\bibitem[Isola et~al.(2017)Isola, Zhu, Zhou, and Efros]{pix2pix2017}
Phillip Isola, Jun-Yan Zhu, Tinghui Zhou, and Alexei~A Efros.
\newblock Image-to-image translation with conditional adversarial networks.
\newblock \emph{CVPR}, 2017.

\bibitem[Jocher et~al.(2023)Jocher, Chaurasia, and Qiu]{yolov8_ultralytics}
Glenn Jocher, Ayush Chaurasia, and Jing Qiu.
\newblock Ultralytics yolov8, 2023.

\bibitem[Kirillov et~al.(2023)Kirillov, Mintun, Ravi, Mao, Rolland, Gustafson, Xiao, Whitehead, Berg, Lo, Dollár, and Girshick]{sam}
Alexander Kirillov, Eric Mintun, Nikhila Ravi, Hanzi Mao, Chloe Rolland, Laura Gustafson, Tete Xiao, Spencer Whitehead, Alexander~C. Berg, Wan-Yen Lo, Piotr Dollár, and Ross Girshick.
\newblock Segment anything, 2023.

\bibitem[Kniaz et~al.(2019)Kniaz, Knyaz, Hlad{\r{u}}vka, Kropatsch, and Mizginov]{thermalgan}
Vladimir~V. Kniaz, Vladimir~A. Knyaz, Ji{\v{r}}{\'i} Hlad{\r{u}}vka, Walter~G. Kropatsch, and Vladimir Mizginov.
\newblock Thermalgan: Multimodal color-to-thermal image translation for person re-identification in multispectral dataset.
\newblock In \emph{Computer Vision -- ECCV 2018 Workshops}, pages 606--624, Cham, 2019. Springer International Publishing.

\bibitem[Kri{\v{s}}to(2016)]{thermal_surveillance1}
Mate Kri{\v{s}}to.
\newblock Review of methods for the surveillance and access control using the thermal imaging system.
\newblock \emph{Review of Innovation and Competitiveness: A Journal of Economic and Social Research}, 2\penalty0 (4):\penalty0 71--91, 2016.

\bibitem[Lahouli et~al.(2018)Lahouli, Haelterman, Chtourou, De~Cubber, and Attia]{litiv}
Ichraf Lahouli, Rob Haelterman, Zied Chtourou, Geert De~Cubber, and Rabah Attia.
\newblock Pedestrian detection and tracking in thermal images from aerial mpeg videos.
\newblock 2018.

\bibitem[Lee et~al.(2023)Lee, Jeon, Cho, and Kim]{eggan}
Dong-Guw Lee, Myung-Hwan Jeon, Younggun Cho, and Ayoung Kim.
\newblock Edge-guided multi-domain rgb-to-tir image translation for training vision tasks with challenging labels.
\newblock \emph{2023 IEEE International Conference on Robotics and Automation (ICRA)}, pages 8291--8298, 2023.

\bibitem[Liao et~al.(2022)Liao, Gao, Li, Wang, and Kwong]{objectdetection2}
Guibiao Liao, Wei Gao, Ge Li, Junle Wang, and Sam Kwong.
\newblock Cross-collaborative fusion-encoder network for robust rgb-thermal salient object detection.
\newblock \emph{IEEE Transactions on Circuits and Systems for Video Technology}, 32\penalty0 (11):\penalty0 7646--7661, 2022.

\bibitem[Liu et~al.(2024)Liu, Zeng, Ren, Li, Zhang, Yang, Jiang, Li, Yang, Su, Zhu, and Zhang]{groundingdino}
Shilong Liu, Zhaoyang Zeng, Tianhe Ren, Feng Li, Hao Zhang, Jie Yang, Qing Jiang, Chunyuan Li, Jianwei Yang, Hang Su, Jun Zhu, and Lei Zhang.
\newblock Grounding dino: Marrying dino with grounded pre-training for open-set object detection, 2024.

\bibitem[Mao et~al.(2024)Mao, Mei, Lu, Liu, Chen, Zhao, and Hu]{PID}
Fangyuan Mao, Jilin Mei, Shun Lu, Fuyang Liu, Liang Chen, Fangzhou Zhao, and Yu Hu.
\newblock Pid: Physics-informed diffusion model for infrared image generation, 2024.

\bibitem[Miethig et~al.(2019)Miethig, Liu, Habibi, and Mohrenschildt]{thermal_auto1}
Ben Miethig, Ash Liu, Saeid Habibi, and Martin~v. Mohrenschildt.
\newblock Leveraging thermal imaging for autonomous driving.
\newblock In \emph{2019 IEEE Transportation Electrification Conference and Expo (ITEC)}, pages 1--5, 2019.

\bibitem[Ozcan and Cetin(2022)]{pedestrian2}
Ahmet Ozcan and Omer Cetin.
\newblock A novel fusion method with thermal and rgb-d sensor data for human detection.
\newblock \emph{IEEE Access}, 10:\penalty0 66831--66843, 2022.

\bibitem[Qiu(2025)]{FLIRalign}
Zona Qiu.
\newblock Flir-align: A repository for flir thermal and rgb image alignment.
\newblock \url{https://github.com/zonaqiu/FLIR-align}, 2025.
\newblock Accessed: 2025-02-20.

\bibitem[Radford et~al.(2021)Radford, Kim, Hallacy, Ramesh, Goh, Agarwal, Sastry, Askell, Mishkin, Clark, Krueger, and Sutskever]{clip}
Alec Radford, Jong~Wook Kim, Chris Hallacy, Aditya Ramesh, Gabriel Goh, Sandhini Agarwal, Girish Sastry, Amanda Askell, Pamela Mishkin, Jack Clark, Gretchen Krueger, and Ilya Sutskever.
\newblock Learning transferable visual models from natural language supervision, 2021.

\bibitem[Ravi et~al.(2024)Ravi, Gabeur, Hu, Hu, Ryali, Ma, Khedr, Rädle, Rolland, Gustafson, Mintun, Pan, Alwala, Carion, Wu, Girshick, Dollár, and Feichtenhofer]{sam2}
Nikhila Ravi, Valentin Gabeur, Yuan-Ting Hu, Ronghang Hu, Chaitanya Ryali, Tengyu Ma, Haitham Khedr, Roman Rädle, Chloe Rolland, Laura Gustafson, Eric Mintun, Junting Pan, Kalyan~Vasudev Alwala, Nicolas Carion, Chao-Yuan Wu, Ross Girshick, Piotr Dollár, and Christoph Feichtenhofer.
\newblock Sam 2: Segment anything in images and videos, 2024.

\bibitem[Ren et~al.(2024)Ren, Liu, Zeng, Lin, Li, Cao, Chen, Huang, Chen, Yan, Zeng, Zhang, Li, Yang, Li, Jiang, and Zhang]{groundedsam}
Tianhe Ren, Shilong Liu, Ailing Zeng, Jing Lin, Kunchang Li, He Cao, Jiayu Chen, Xinyu Huang, Yukang Chen, Feng Yan, Zhaoyang Zeng, Hao Zhang, Feng Li, Jie Yang, Hongyang Li, Qing Jiang, and Lei Zhang.
\newblock Grounded sam: Assembling open-world models for diverse visual tasks, 2024.

\bibitem[Rombach et~al.(2022)Rombach, Blattmann, Lorenz, Esser, and Ommer]{stable_diffusion}
Robin Rombach, Andreas Blattmann, Dominik Lorenz, Patrick Esser, and Bj{\"o}rn Ommer.
\newblock High-resolution image synthesis with latent diffusion models.
\newblock In \emph{Proceedings of the IEEE/CVF conference on computer vision and pattern recognition}, pages 10684--10695, 2022.

\bibitem[Sun et~al.(2020)Sun, Zuo, Yun, Wang, and Liu]{segmentation1}
Yuxiang Sun, Weixun Zuo, Peng Yun, Hengli Wang, and Ming Liu.
\newblock Fuseseg: Semantic segmentation of urban scenes based on rgb and thermal data fusion.
\newblock \emph{IEEE Transactions on Automation Science and Engineering}, 18\penalty0 (3):\penalty0 1000--1011, 2020.

\bibitem[Takumi et~al.(2017)Takumi, Watanabe, Ha, Tejero-De-Pablos, Ushiku, and Harada]{mfnet}
Karasawa Takumi, Kohei Watanabe, Qishen Ha, Antonio Tejero-De-Pablos, Yoshitaka Ushiku, and Tatsuya Harada.
\newblock Multispectral object detection for autonomous vehicles.
\newblock In \emph{Proceedings of the on Thematic Workshops of ACM Multimedia 2017}, page 35–43, New York, NY, USA, 2017. Association for Computing Machinery.

\bibitem[Torresan et~al.(2004)Torresan, Turgeon, Ibarra-Castanedo, Hebert, and Maldague]{thermal_surveillance2}
Helene Torresan, Benoit Turgeon, Clemente Ibarra-Castanedo, Patrick Hebert, and Xavier~P Maldague.
\newblock Advanced surveillance systems: combining video and thermal imagery for pedestrian detection.
\newblock In \emph{Thermosense XXVI}, pages 506--515. SPIE, 2004.

\bibitem[Wan et~al.(2024)Wan, Wang, Yong, Zhang, Stepputtis, Sycara, and Xie]{sigma}
Zifu Wan, Yuhao Wang, Silong Yong, Pingping Zhang, Simon Stepputtis, Katia Sycara, and Yaqi Xie.
\newblock Sigma: Siamese mamba network for multi-modal semantic segmentation.
\newblock \emph{arXiv preprint arXiv:2404.04256}, 2024.

\bibitem[Xie et~al.(2021)Xie, Wang, Yu, Anandkumar, Alvarez, and Luo]{segformer}
Enze Xie, Wenhai Wang, Zhiding Yu, Anima Anandkumar, Jose~M. Alvarez, and Ping Luo.
\newblock Segformer: simple and efficient design for semantic segmentation with transformers.
\newblock Red Hook, NY, USA, 2021. Curran Associates Inc.

\bibitem[Zhang et~al.(2023)Zhang, Huang, Ma, Li, Luo, Xie, Qin, Luo, Li, Liu, Guo, and Zhang]{ram}
Youcai Zhang, Xinyu Huang, Jinyu Ma, Zhaoyang Li, Zhaochuan Luo, Yanchun Xie, Yuzhuo Qin, Tong Luo, Yaqian Li, Shilong Liu, Yandong Guo, and Lei Zhang.
\newblock Recognize anything: A strong image tagging model, 2023.

\bibitem[Zhou et~al.(2021)Zhou, Guo, Lei, Yu, and Hwang]{objectdetection1}
Wujie Zhou, Qinling Guo, Jingsheng Lei, Lu Yu, and Jenq-Neng Hwang.
\newblock Ecffnet: Effective and consistent feature fusion network for rgb-t salient object detection.
\newblock \emph{IEEE Transactions on Circuits and Systems for Video Technology}, 32\penalty0 (3):\penalty0 1224--1235, 2021.

\bibitem[Zhu et~al.(2017)Zhu, Park, Isola, and Efros]{cyclegan}
Jun-Yan Zhu, Taesung Park, Phillip Isola, and Alexei~A Efros.
\newblock Unpaired image-to-image translation using cycle-consistent adversarial networks.
\newblock In \emph{Proceedings of the IEEE international conference on computer vision}, pages 2223--2232, 2017.

\bibitem[Özkanoğlu and Ozer(2022)]{infragan}
Mehmet~Akif Özkanoğlu and Sedat Ozer.
\newblock Infragan: A gan architecture to transfer visible images to infrared domain.
\newblock \emph{Pattern Recognition Letters}, 155:\penalty0 69--76, 2022.

\end{thebibliography}
}

\end{document}